\documentclass[onecolumn,11pt]{article}
\usepackage[top=1in, bottom=1in, left=1in, right=1in]{geometry}
\usepackage{amsmath,amsfonts,amscd,amssymb,amsthm}
\usepackage{graphicx}
\usepackage{epstopdf}
\usepackage{overpic}
\usepackage{cancel}
\usepackage{rotating}
\usepackage{url}
\usepackage{caption}
\usepackage{color}
\usepackage{rotating}
\usepackage{multirow}
\usepackage{wrapfig}
\usepackage{mathtools}
\usepackage{subeqnarray}
\usepackage{setspace}
\usepackage{palatino} 
\usepackage[numbers,sort&compress]{natbib}
\usepackage{bm}
\usepackage{physics}

\theoremstyle{remark}

\newtheorem{hypothesis}{Hypothesis}
\newcommand*{\vertbar}{\rule[-1ex]{0.5pt}{2.5ex}}

\usepackage[bottom,flushmargin,hang,multiple]{footmisc}
\usepackage{lipsum}
\newcommand\blfootnote[1]{%
  \begingroup
  \renewcommand\thefootnote{}\footnote{#1}%
  \addtocounter{footnote}{-1}%
  \endgroup
}

\definecolor{header1}{cmyk}{0,0,0,1}

\DeclareMathOperator*{\argmin}{arg\rm{}min}

\title{\vspace{-.25in}{\LARGE \textbf{A Data-Driven Buckingham Pi Analysis\\ for Dimensionally Consistent Learning}}\vspace{-.15in}}
\title{\vspace{-.55in}{\LARGE \textbf{Buckingham Pi Analysis for Dimensionally Consistent Learning}}\vspace{-.15in}}
\title{\vspace{-.55in}{\LARGE \textbf{Buckingham Pi for Dimensionally Consistent Learning}}\vspace{-.15in}}
\title{\vspace{-.55in}{\LARGE \textbf{Constraining Machine Learning for Dimensional Consistency with Buckingham Pi}}\vspace{-.15in}}
\title{\vspace{-.25in}{\LARGE \textbf{Dimensionally Consistent Learning with Buckingham Pi}}\vspace{-.15in}}

\author{\normalsize{Joseph Bakarji$^{1}$, Jared Callaham$^1$, Steven L. Brunton$^1$, J. Nathan Kutz$^2$}\\
\footnotesize{$^1$ Department of Mechanical Engineering, University of Washington, Seattle, WA 98195, United States}\\
\footnotesize{$^2$ Department of Applied Mathematics, University of Washington, Seattle, WA 98195, United States\vspace{-.1in}}
}
\date{}
\graphicspath{/figures}
\begin{document}
\maketitle

\blfootnote{$^*$ Corresponding author (jbakarji@uw.edu).}
\vspace{-.2in}
\begin{abstract}
    In the absence of governing equations, dimensional analysis is a robust technique for extracting insights and finding symmetries in physical systems. 
    Given measurement variables and parameters, the Buckingham Pi theorem provides a procedure for finding a set of dimensionless groups that spans the solution space, although this set is not unique. 
    We propose an automated approach using the symmetric and self-similar structure of available measurement data to discover the dimensionless groups that best collapse this data to a lower dimensional space according to an optimal fit. 
    We develop three data-driven techniques that use the Buckingham Pi theorem as a constraint: (i) a constrained optimization problem with a non-parametric input-output fitting function, (ii) a deep learning algorithm (BuckiNet) that projects the input parameter space to a lower dimension in the first layer, and (iii) a technique based on sparse identification of nonlinear dynamics (SINDy) to discover dimensionless equations whose coefficients parameterize the dynamics.
    We explore the accuracy, robustness and computational complexity of these methods as applied to three example problems: a bead on a rotating hoop, a laminar boundary layer, and Rayleigh-B\'enard convection.  \\

\end{abstract}

\section{Introduction}

Dimensional analysis is based on the simple idea that physical laws do not depend on the units of measurements.  
As a consequence, any function that expresses a physical law has the fundamental property of so-called {\em generalized homogeneity}~\cite{barenblatt1996scaling} and does not depend on the observer.  
Although such concepts of dimensional analysis go back to the time of Newton and Galileo~\cite{sterrett2017physically}, it was formalized mathematically by the pioneering contributions of Edgar Buckingham in 1914~\cite{buckingham1914physically}.   
Specifically, Buckingham proposed a principled method for extracting the most general form of physical equations by simple dimensional considerations of the seven fundamental units of measurement:  length (meter), mass (kg), time (seconds), electric current (Ampere), temperature (Kelvin), amount of substance (mole), and luminous intensity (candela).   
From electromagnetism to gravitation, measurements can be directly related to the seven fundamental units, i.e. force is measured in Newtons which is kg$\cdot$m/s$^2$ and electric charge by the volt which is kg$\cdot$m$^2$/(s$^3$$\cdot$A).  
The resulting Buckingham Pi theorem was originally contextualized in terms of physically similar systems~\cite{buckingham1914physically}, or groups of parameters that related similar physics.  
In modern times, rapid advancements in measurement and sensor technologies have produced data of diverse quantities at almost any spatial and temporal scale.  
This has engendered a renewed consideration of the Buckingham Pi theorem in the context of data-driven modeling~\cite{del2019lurking,jofre2020data,fukami2021robust,xie2021data}.  
Specifically, as shown here, physics-informed machine learning algorithms can be constructed to automate the principled approach advocated by Buckingham for the discovery of the most general and parsimonious form of physical equations possible from measurements alone.

Buckingham Pi theory was critically important in the pre-computer era.  Indeed, the discovery of  key dimensionless quantities was considered a fundamental result, as it often uncovered the self-similarity structure of the solution along with parametric dependencies and symmetries.  In fluid dynamics alone, for instance, there are numerous named dimensionless parameters that are discovered through the Buckingham Pi reduction, including the Reynolds, Prandtl, Froude, Weber, and Mach numbers. 
Scaling and dimensional analysis continue to provide a robust starting point for understanding both complex and basic physical phenomena, like the physics of wrinkling~\cite{Cerda2003prl}, chaotic patterns in Rayleigh-B\'enard convection~\cite{Morris1993prl}, the structure of drops falling from a faucet \cite{Shi1994science}, spontaneous pattern formation by self-assembly~\cite{Grzybowski2000nature}, and osmotic spreading of biofilms~\cite{Seminara2012pnas}.
In practice, many dimensionless groups naturally arise from ratios of domain averaged terms in physically meaningful equations that are required to be dimensionally homogeneous.   When equations are non-dimensionalized, dimensionless terms that are small on average relative to others can be treated as perturbations, thus providing insights into the qualitative structure of the solution and simplifying the problem of finding it. More importantly, Buckingham Pi provides the potential for an {\em order parameter} description~\cite{Cross1993} that determines the qualitative structure of the solution and its bifurcations without recourse to the full-state equations~\cite{callaham2021learning}.
These order parameters form the basis for casting problems into normal forms that reveal physical symmetries and can potentially be solved analytically~\cite{guckenheimer_holmes, Yair2017pnas, kalia2021learning}.
Even when governing equations are unknown, Buckingham Pi establishes a dimensionally consistent relationship between the input parameters/variables and output predictions which help constrain models and prevent over-fitting.

The rise of scientific computing in the 1980s made it possible to numerically integrate highly nonlinear ordinary and partial differential equations, without the need for analytic simplifications provided by dimensional analysis.   
More recently, machine learning algorithms have shown promise in discovering scientific laws~\cite{schmidt2009distilling}, differential equations~\cite{brunton2016discovering, rudy2017data, lu2021learning}, and deep network input-output function approximations~\cite{raissi2019physics, karniadakis2021physics, Noe2019science} from simulation and experimental data alone, with considerable progress in the field of fluid mechanics~\cite{Brenner2019prf,Duraisamy2019arfm,Brunton2020arfm,Sonnewald2021}.
Particularly, unsupervised learning techniques have recently made significant progress in distilling physical data into interpretable dynamical models that automate the scientific process \cite{Kaiser2021, Sonnewald2019ess, wu2018deep, Champion2019pnas, bakarji2022discovering}.
However, these methods do not take into account the units of their training data, which can significantly constrain the hypothesis class to physically meaningful solutions. 
{Principal component analysis} (PCA), which is a leading data analysis method~\cite{Brunton2019book}, attempts to circumvent the idea that physical laws do not depend on the units of measurements by taking each measurement variable and setting its mean to zero and its variance to unity. 
However, there are more sophisticated methods for explicitly considering measurement units.   
Active subspaces provides a principled algorithmic method for extracting dimensionless numbers from experimental and simulation data~\cite{constantine2017data}, having successfully been demonstrated to discover dimensionless numbers in particle-laden turbulent flows~\cite{jofre2020data}.
In a related line of work, statistical null-hypothesis testing has been used alongside the Buckingham Pi theorem to find hidden variables in dimensional experimental data~\cite{del2019lurking}.
Dimensional analysis has also been used for a physics-inspired symbolic regression algorithm that discovers physics formulas from data~\cite{udrescu2020ai} and to improve neural network models~\cite{gunaratnam2003improving}. 
Data-driven dimensional analysis has also been applied to model turbulence data~\cite{fukami2021robust}. 
Machine learning algorithms that use sparse identification of differential equation to discover dimensionless groups and scaling laws that best collapse experimental data have also been recently proposed~\cite{xie2021data}.

\begin{figure}[t]
\begin{center}
\includegraphics[width=.85\textwidth]{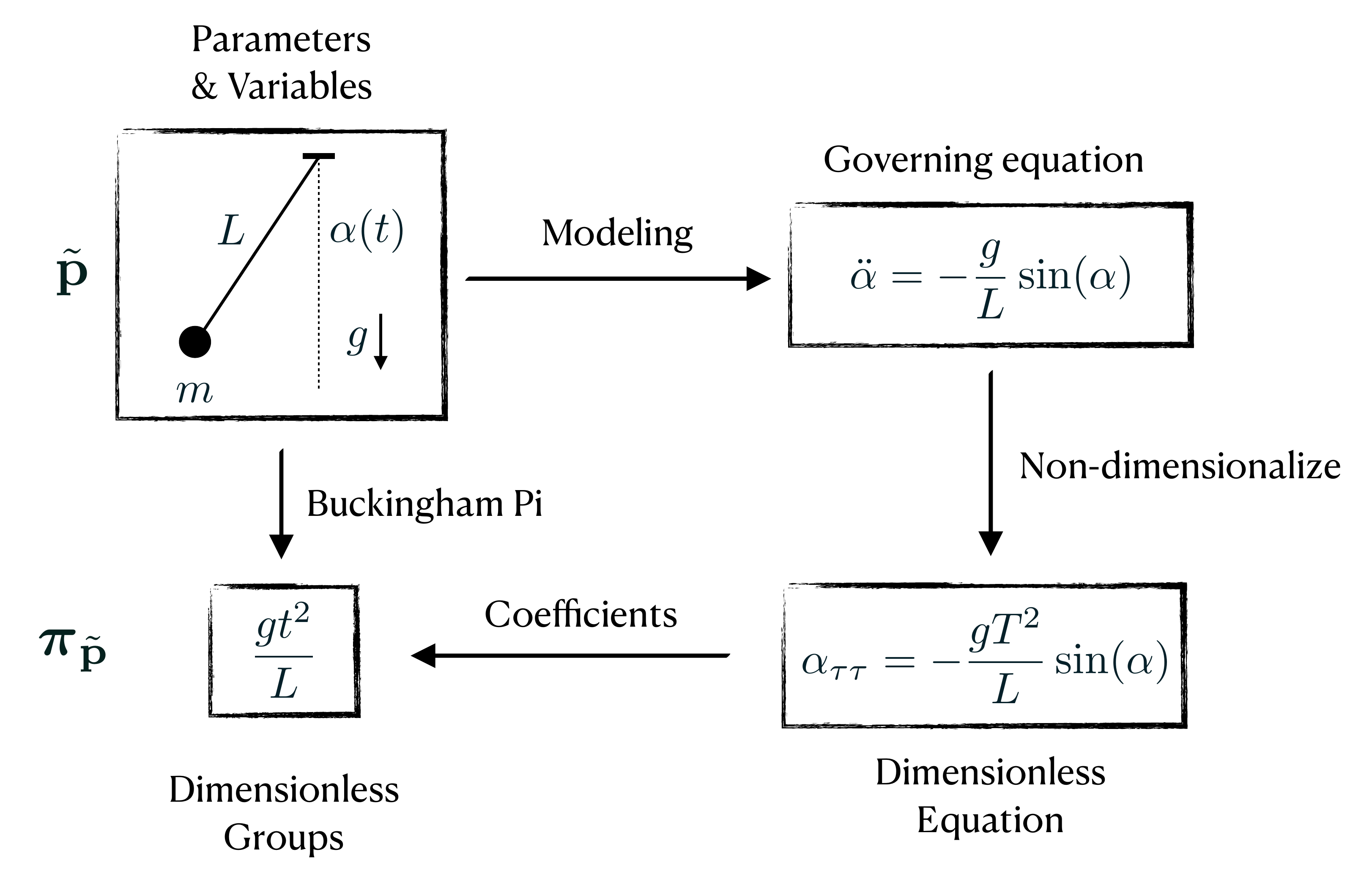}
\caption{Two approaches for non-dimensionalization. In the absence of known governing equations, scientists use the Buckingham Pi theorem to infer dimensionless groups $\bm \pi_{\tilde{\mathbf p}}$ from units of input variables and parameters. When the equation is known, normalizing variables and parameters by system-specific constants and/or averages gives rise to dimensionless groups as equation coefficients.}
\label{fig:approaches}
\end{center}
\end{figure}

In this study, we address the problem of automatic discovery of dimensionless groups from available experimental or simulation data using the Buckingham Pi theorem as a constraint.  
Importantly, although the Buckingham Pi theorem provides a step-by-step method for finding dimensionless groups directly from the input variables and parameters, its solution is not unique. 
Thus it relies on intuition and experience with the physical problem at hand. 
If the underlying equations are available, finding dimensionless groups is constrained by the form of the equation and they are computed by dividing the \emph{system average} magnitude of every term by a reference term. 
Even then, the problem of determining the relevant dimensionless numbers depends on experience.  
Our aim is to leverage symmetries in the data to inform the algorithm of which set of dimensionless groups control the behavior (e.g. the Reynolds number controls the turbulence of a flow-field). 
We develop three methods that find the most physically meaningful Pi groups (See Fig.~\ref{fig:approaches}), assumed to be the most predictive of the dependent variables. 
The first method is a constrained optimization that fits independent variable and parameter inputs to predictions, while satisfying the Buckinham Pi theorem as a hard constraint. 
The second method uses a Buckingham Pi (nullspace) constraint on the first hidden layer of a deep network that fits input parameters to output predictions. 
And the third method uses SINDy~\cite{brunton2016discovering, rudy2017data} to constrain available dimensionless groups to be coefficients in a sparse differential equation. 
The latter method has the added benefit of discovering a parametric (dimensionless) differential equation that describes the data, thus simultaneously generalizing the SINDy method. 
We apply these methods to three problems: the rotating hoop, the Blasius boundary layer, and the Rayleigh-B\'ernard problem.

\section{Buckingham Pi Problem Formulation}\label{sec:problem-statement}

Let $\mathbf{q}= f(\mathbf p)$ be a mapping where $\mathbf p \in \mathbb R^n$ is a vector of input measurement parameters and variables, and $f$ is a function that maps $\mathbf p$ to a set of dependent quantities of interest $\mathbf{q} \in \mathbb R^k$. One generally seeks a predictor $f$ that maximizes the size of the set of possible measurable parameters and variables $\mathcal P$, for which the prediction error $\varepsilon = \| f(\mathbf p) - \mathbf{q} \|$ is minimized, with $\mathbf p, \mathbf q \in \mathcal P$. The function $f$ is typically the solution of an initial and/or boundary value problem with known variables and parameters $\mathbf p$ and unknown solution $\mathbf{q}$, all of which have physical dimensions [units]. 

The Buckingham Pi theorem states that there exists a set of dimensionless quantities $\bm \pi = (\bm \pi_p, \bm \pi_q)$, also known as $\bm \pi-$groups, such that the dimensionless input parameters/variables $\bm \pi_p \in \mathbb R^{n'}$ span the full dimensionless solution space $\bm \pi_q \in \mathbb R^{k'}$, with $n' \le n$ and $k' \le k$~\cite{buckingham1914physically}. Furthermore, the theorem determines the value of $d' = n'+k' $ given the units of the measurement variables and parameters. 

A probabilistic corollary of the Buckingham Pi theorem is that the input-output data $(\mathbf p, \mathbf q)$ can be projected on a lower dimensional space with basis $(\bm \pi_p, \bm \pi_q)$ without compromising the prediction error $\varepsilon$, such that, 
\begin{equation}
\left\| \mathbf{q} - f(\mathbf p) \right\| < \varepsilon \quad \rightarrow \quad \left\| \bm \pi_q - \psi(\bm \pi_p) \right\| < \varepsilon,
\end{equation}
where $\psi$ is an unknown function that can be approximated from available data. 
However, the Buckingham Pi theorem does not provide any information on the properties of $\psi$ or its relationship with the Pi groups. The theorem generally assumes $\mathbf p$, $\mathbf q$, and $f$ to be deterministic without considering the optimality of a dimensionless basis with the generalization properties of the mapping $f$.

In a data-driven statistical setting, we assume that the inputs $\mathbf p$ and outputs $\mathbf q$ are sampled from a distribution function $F_{\mathbf p \mathbf q}(P, Q)$, and that $f$ is an optimal mapping in a given class of function. In this context, we generally seek the most physically meaningful transformation $(\mathbf{p}, \mathbf q) \rightarrow (\bm \pi_p, \bm \pi_q)$ that provides the best fit  $\bm \pi_q = \psi(\bm \pi_p)$, assuming $\psi$ to be an optimal predictor over an pre-determined hypothesis class $\mathcal H$. Accordingly, we posit the following hypothesis

\begin{hypothesis}\label{main-hyp}
Given a set of input and output measurement pairs $\tilde{\mathbf p} = (\mathbf p, \mathbf q)$, the most physically meaningful dimensionless basis $\bm \pi^*$ is the optimal coordinate transformation $\tilde{\mathbf p} \rightarrow \bm \pi^*$ that satisfies the minimization problem
\begin{equation}\label{eq:optim-thrm}
\bm \pi^* = \underset{\bm \pi}{\argmin} \left( \underset{\psi}{\min} \left\| \bm \pi_q - \psi(\bm \pi_p)\right\|^2_2 \right),
\end{equation}
where \emph{physical meaningfulness} is defined as the dimensionless group's ability to simplify an equation in relevant regimes and provide a scaling that collapses the input-output solution space to a lower dimension. 
\end{hypothesis}

Note that defining \emph{physical meaningfulness} often depends on the historical and scientific context, which makes the above hypothesis subject to interpretation. In some cases, theoretical and experimental tradition narrows a field of study down to a set of well-known dimensionless numbers with which known regimes and behaviors are associated (e.g. Reynolds number quantifying turbulence in fluid mechanics). Particularly, when analytical models and equations are available, dimensionless numbers that arise naturally from the relative measure of different terms are physically significant because they quantify changes in the qualitative nature of the solution. Hypothesis~\ref{main-hyp} proposes a new approach for defining \emph{physical meaningfulness} by first providing evidence for its validity in Sec.\ref{sec:methods}. The purpose of the proposed methods is to discover dimensionless groups in contexts where data is available but analytical models are not known.


This hypothesis is further constrained by the functional relationship between $\bm \pi$ and $\bm \tilde{\mathbf p}$ given by 
\begin{equation}\label{eq:ptopi}
\pi_j = \prod_{i=1}^{d} \tilde{p}_i^{\Phi_{ij}},
\end{equation}
where $\Phi_{ij}$ are unknown parameters that are chosen to make $\pi_j$ dimensionless. 

Let $\Omega()$ be a function that maps a measurable quantity $\tilde p_i$ to a vector containing the powers of its basic dimensions (i.e. mass $M$, length $L$, time $T$, etc.). For example, if one chooses $[M, L, T]$ as the basic dimensions, then the units of a force are $[F] = ML/T^2$, and $\Omega(F) = [1, 1, -2]^T$. Let $\phi()$ be a function that maps a dimensionless group to a vector that corresponds to the powers of the input parameters $\mathbf p$ such that $\phi(\pi_j) = [\Phi_{1j}, \Phi_{2j}, \ldots, \Phi_{dj}]^T$, where $d=n+k$.

The Buckingham Pi theorem~\cite{buckingham1914physically} constrains $\phi(\pi_i)$ to be in the null-space of the units matrix
\begin{equation}
\bm D = [\bm D_p, \bm D_q] =
\left[
  \begin{array}{ccccccc}
    \vertbar & \vertbar &        & \vertbar  & \vertbar & & \vertbar  \\
    \Omega( p_1)    & \Omega( p_2)    & \ldots & \Omega( p_n) & \Omega( q_1)    & \ldots & \Omega( q_k)    \\
    \vertbar & \vertbar &        & \vertbar  & \vertbar & & \vertbar 
  \end{array}
\right],
\end{equation}
such that
\begin{equation}\label{eq:nulleqn}
\bm D \phi(\pi_i) = \mathbf 0, \quad \forall i \in \{1,\ldots, d\}.
\end{equation}
Here, we assume that the number of dimensional and dimensionless predictions are equal, i.e. $|\bm \pi_q| = |\mathbf q| = k$, which is often the case, and we later propose methods for relaxing this assumption. 
%
The Buckingham Pi theorem determines the number of Pi-groups to be $d' = d - \text{rank}(\bm D)$. The main shortcoming of the theorem is its inability to provide a unique set of $d'$ dimensionless numbers. In practice, additional heuristic constraints are required to solve for the unknowns $s_{ij}$.

\section{Methods}
\label{sec:methods}

We present three methods for discovering dimensionless groups from data. The main features that differentiate them are 
\begin{enumerate}
    \item Imposing hard, soft, or no constraints on the null-space of $\bm D$ to ensure that the Pi-groups are dimensionless according to equation~\eqref{eq:nulleqn}.
    \item The type of input-output function approximation $\psi$: e.g. neural network, non-parametric function, or differential equation.
\end{enumerate}

Let $\{\tilde{\mathbf{p}}^{(i)}\}_{i=1}^m \equiv \{(\mathbf p^{(i)}, \mathbf q^{(i)})\}_{i=1}^m$ be a set of $m$ measurement pairs of the inputs vector $\mathbf p^{(i)} \in {\mathbb R}^n$ and their corresponding output predictions $\mathbf q^{(i)} \in {\mathbb R}^k$, with index $i$.
We define the input parameter matrix $\boldsymbol{P} \in {\mathbb R}^{m \times n}$ and the output prediction matrix $\boldsymbol{Q} \in {\mathbb R}^{m \times k}$ as the row-wise concatenation of all measurements $\mathbf p^{{(i)}}$ and $\mathbf q^{{(i)}}$ respectively. We also define $\tilde{\boldsymbol{P}} = [\bm P, \bm Q] \in {\mathbb R}^{m \times d}$ as the column-wise concatenation of $\boldsymbol{P}$ and $\boldsymbol{Q}$. 
The Pi-groups powers matrix is given by
\begin{equation}
\bm \Phi = 
\left[
  \begin{array}{cccc}
    \vertbar & \vertbar &        & \vertbar \\
    \phi(\pi_1)    & \phi(\pi_2)    & \ldots & \phi(\pi_{d'})    \\
    \vertbar & \vertbar &        & \vertbar 
  \end{array}
\right] \in {\mathbb R}^{d \times {d'}},
\end{equation}
such the equation $\bm D \bm \Phi = \mathbf{0}$ is satisfied according to Eq.\eqref{eq:nulleqn}. Similarly, we define $\bm \Phi_p \in {\mathbb R}^{n\times n'}$ to contain the $n'$ dimensionless groups corresponding to the $n$ input parameters/variables $\mathbf p$. Equation~\eqref{eq:ptopi} can be written as $\pi_i = \exp \left\{ \sum_{j=1}^d \Phi_{ij}\log(\tilde p_j) \right\}$, corresponding to the matrix form
\begin{equation}\label{eq:ptopi-mat}
    \boldsymbol{\Pi} = \exp \left( \log(\boldsymbol{\tilde{P}})  \boldsymbol{\Phi}  \right),
\end{equation}
where each row in $\boldsymbol{\Pi} \in \mathbb R^{m\times d'}$ corresponds to the values of the dimensionless groups $\bm \pi^{(i)}$ for a given experiment $i$, and the exponential and logarithm are both taken element-wise. We also define the matrix $\bm \Pi_q \in {\mathbb R}^{m \times k'}$ to be the data matrix of the output Pi-groups $\bm \pi_q$ and $\bm \Pi_p \in {\mathbb R}^{m \times n'}$ to be the data matrix of the input Pi-groups $\bm \pi_p$, such that $\bm \Pi = [\bm \Pi_p, \bm \Pi_q]$.

For example, consider the pendulum shown in Fig.~\ref{fig:approaches}.
If we define the inputs to be gravity $g$, mass $m$, length $L$, and time $t$, and the output to be the angle $\alpha$, then the dimensional input and output parameter vectors $\mathbf p$ and $\mathbf q$ are
\begin{equation}
    \mathbf p =
    \begin{bmatrix}
    g & m & L & t
    \end{bmatrix}^T, \qquad
    \mathbf q =
    \begin{bmatrix}  \alpha \end{bmatrix}.
\end{equation}
The corresponding units matrices $\bm D_p$ and $\bm D_q$ constructed with basic dimensions mass, length, and time are
\begin{subequations}
\begin{align}
    \bm D_p &=
    \left[
  \begin{array}{ccccc}
    \vertbar & \vertbar &  \vertbar  & \vertbar  \\
    \Omega( g )    & \Omega( m )  & \Omega( L )    &  \Omega( t )    \\
    \vertbar &  \vertbar     & \vertbar  & \vertbar
  \end{array}
\right]
  = \left[
  \begin{array}{ccccc}
    0 & 1  &  0  & 0   \\
    1  & 0  & 1 & 0  \\
    -2 & 0 & 0 & 1
  \end{array}
\right] \\
 \bm D_q &= \left[
  \begin{array}{c}
    \vertbar \\
    \Omega(\alpha)   \\
    \vertbar
  \end{array}
\right] = \left[
  \begin{array}{c}
    0 \\
    0   \\
    0 
  \end{array}
\right].
\end{align}
\end{subequations}
Note that the angle $\alpha$ is already dimensionless, so $\Omega(\alpha)$ is the zero vector.
A possible $\bm \Phi$ matrix consisting of columns spanning the nullspace of $\bm D = [\bm D_p, \bm D_q]$ is
\begin{equation}
    \bm \Phi = \left[
  \begin{array}{cc}
    1 & 0    \\
    0 & 0  \\
    -1 & 0  \\
    2 & 0 \\
    0 & 1
  \end{array}
\right]
\quad
\text{s.t.}
\quad
\bm \Phi_p = \left[
  \begin{array}{c}
    1    \\
    0  \\
   -1  \\
    2 \\
    0
  \end{array}
\right], 
\quad
\bm \Phi_q = \left[
  \begin{array}{c}
    0    \\
    0   \\
    0   \\
    0  \\
    1 
  \end{array}
\right].
\end{equation}
Then $\pi_q = \alpha$ and $\pi_p = gt^2/L$, so that the input/output relationship is some function $\alpha = \psi(gt^2/L)$. We solve this example using a data-driven method in Sec.~\ref{sec:results-pendulum}.

In this case we know from basic mechanics that $\psi() \equiv \sin()$ for small $\alpha$, but for more complicated problems it may not be obvious what the appropriate $\bm \Phi$ and $\psi$ are.
This section introduces three methods to simultaneously identify the nondimensionalization $\bm \Phi$ and an approximation of $\psi$ from a set of experimental or simulation data.
In this data-driven context, \eqref{eq:ptopi-mat} can be combined with the nullspace constraint, $\bm D \bm \Phi = \mathbf{0}$, to optimize $\psi$ over a pre-defined hypothesis class given the measurement data.

\subsection{Constrained optimization}
\label{sec:constrained-opt}
Equation~\eqref{eq:optim-thrm} can be cast as a constrained optimization problem, with a fitting function $\psi$ that can be either parameter or non-parametric. The input Pi-groups powers matrix $\bm \Phi_p$ maps the input parameters/variables to the dimensionless groups through equation~\eqref{eq:ptopi-mat}, and the constraint of the optimization is given by equation~\eqref{eq:nulleqn}. The choice of the hypothesis class of $\psi$ is assumed to be arbitrary in this formulation, notwithstanding its effect on the accuracy of the results and the success of the method.

In this setup, we assume that the outputs, $\mathbf q$, can be non-dimensionalized by known constants of motion (i.e. $\bm \Pi_q$ is known). Accordingly, the resulting optimization problem is given by
\begin{equation}
\label{eq:constrained-opt}
    \bm{\check{\Phi}_p} = \argmin_{\boldsymbol{\Phi}_p} \left\|  \boldsymbol{\Pi}_q - \psi(\exp \left( \log(\boldsymbol{P})  \boldsymbol{\Phi}_p  \right)) \right\|_2 + \lambda_1 \left\| \boldsymbol{\Phi}_p \right\|_1 + \lambda_2 \left\| \boldsymbol{\Phi}_p \right\|_2, \quad \text{s.t.} \quad \boldsymbol{D}_p \boldsymbol{\Phi}_p = \mathbf{0},
\end{equation}
where the $\ell_1$ regularization enforces sparsity (typical in dimensionless numbers) and the $\ell_2$ regularization encourages smaller powers. 
An example application of this approach that uses kernel ridge regression as a non-parametric approximation of $\psi$ is given in Sec.~\ref{sec:results-blasius}.

\subsection{BuckiNet: non-dimensionalization with a neural network}
\begin{figure}[t]
\begin{center}
\includegraphics[width=\textwidth]{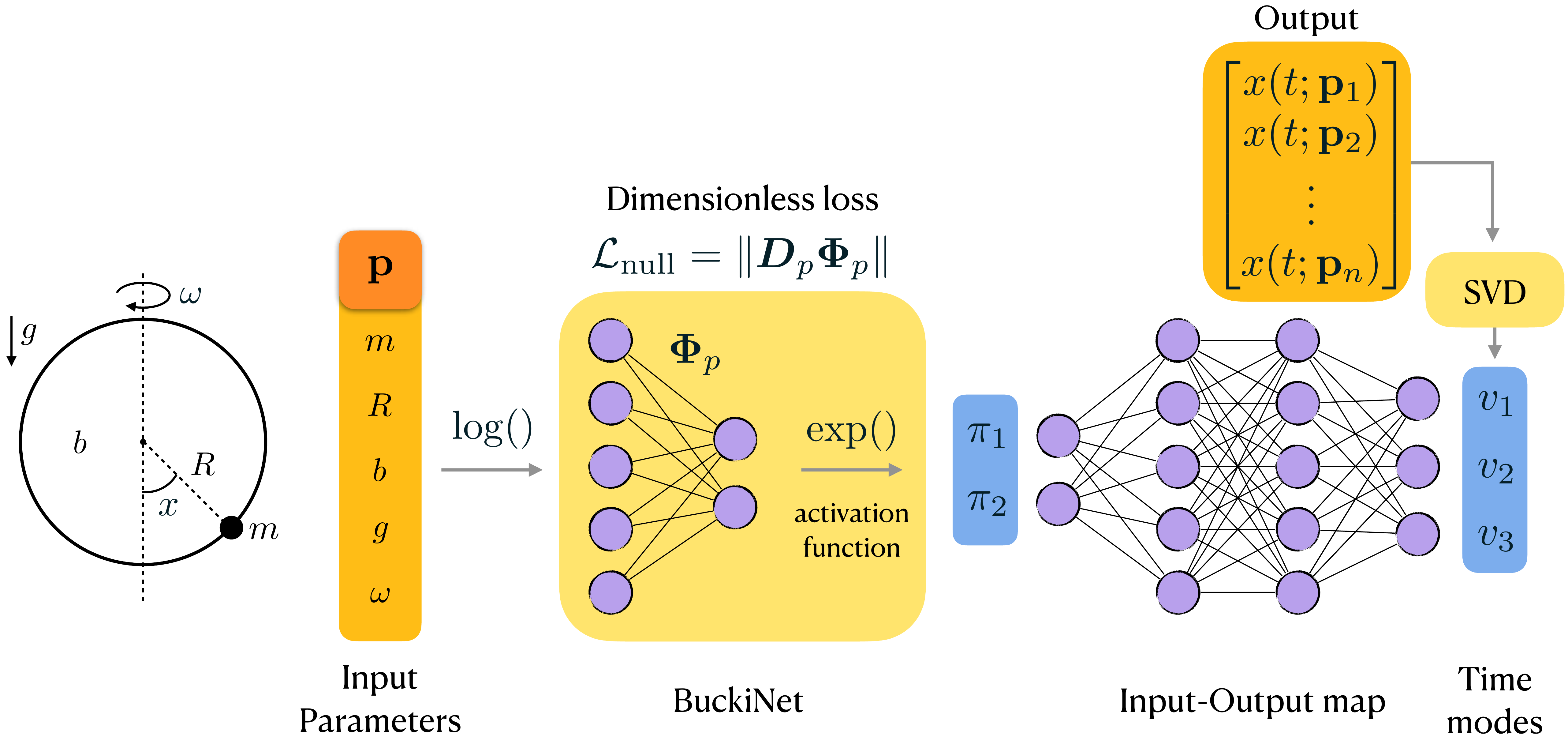}
\caption{Illustration of the BuckiNet layer for the rotating hoop problem (described in section~\ref{sec:rotating-hoop}). The dimensionless loss imposes a soft Buckingham Pi constraint from equation \eqref{eq:nulleqn}) and the BuckiNet layer satisfies equation \eqref{eq:ptopi}. In this example, the output is given by the three dominant time modes $\mathbf v$ of the output $x(t)$ rather than taking the time $t$ as an input.}
\label{fig:neural-nondim}
\end{center}
\end{figure}

The unknown function $\psi$ is generally nonlinear and can be high dimensional. Given input-output data, in the absence of a governing equation, a deep neural network is a natural candidate for approximating $\psi$. 
Equation~\eqref{eq:ptopi-mat} can be naturally integrated in a deep learning architecture by first applying a $\log()$ transform to the inputs $\mathbf p$, then using an exponential activation function at the output of the first layer. This makes $\mathbf \Phi_p$ the fitting weights of the first layer, as shown in Fig.~\ref{fig:neural-nondim}. We call this layer the BuckiNet. 
When negative input data is given, it has to be pre-processed so that the initial $\log()$ operation becomes possible. In most cases, a simple shifting of the data to the positive domain is sufficient.

This technique offers many advantages. First, the BuckiNet layer implicitly performs an unsupervised dimensionality reduction without any adjustment to the overall architecture of the deep network. This results in better generalization properties and a faster optimization thanks to a loss-less reduction in the number fitting parameters. The number of nodes in the first hidden layer is $n'$ as determined by the Buckingham Pi theorem. Second, the optimal weights of the first layer, $\bm \Phi_p$, correspond to the powers of dimensionless numbers that are most predictive of the solution. Third, the BuckiNet can be easily added to deep learning algorithms that fit input-output measurement data, with a few lines of code on Tensorflow or PyTorch. 

The BuckiNet architecture on its own does not guarantee a dimensionless combination of the input parameters. Whether the network discovers a dimensionless basis without any constraint is discussed in Sec.\ref{sec:results-pendulum}. To explicitly account for the constraint in equation~\eqref{eq:nulleqn}, we add a null-space loss on $\bm \Phi_p$
\begin{equation}
\mathcal L_\text{null} = \left\| \bm D_p \bm \Phi_p \right\|^2_2,
\end{equation}
in the loss function. Therefore, the total loss is given by
\begin{equation}
    \mathcal L = \left\|  \boldsymbol{\Pi}_q - \psi(\exp \left( \log(\boldsymbol{P})  \boldsymbol{\Phi}_p  \right)) \right\|_2 + \lambda \| \bm D_p \bm \Phi_p \|_2 + \text{reg}.
\end{equation}
In contrast to the constrained optimization problem proposed in the previous section, this method minimizes the null-space according to equation \eqref{eq:nulleqn} but does not satisfy it exactly. However, this proves to be sufficient for finding Pi-groups with sparse and approximately rational exponents as shown in Sec.\ref{sec:rotating-hoop}. The regularization term includes the $L_2$ loss $\| \bm \Phi_p \|_2$ that promotes dimensionless groups with lower powers, and the $L_1$ loss $\| \bm \Phi_p \|_1$ that promotes simple groups that have as little input variables/parameters as possible.

\subsection{Sparse identification of dimensionless equations}
\label{sec:methods-sindy}

The previous two methods addressed the problem of data-driven non-dimensionalization by simultaneously optimizing for the fitting function $\psi$ and the dimensionless groups $\bm \pi$. 
In light of the fact that $\psi$ is usually the solution of a differential equation, we propose casting the problem as a sparse identification of differential equations (SINDy)~\cite{brunton2016discovering,rudy2017data} with candidate dimensionless groups as coefficients.
This constrains the dimensionless groups to be physically meaningful according to their associated differential operators that act on the unknown output variables, which is also the intuition often used in classical non-dimensionalization.

In the following formulation, we assume that the time $t$ is the only dependent variable, and non-dimensionalize it separately from the rest of the input parameters. 
However, the method can be generalized to any number of dependent variables.
For a dimensionless quantity of interest $\boldsymbol{\pi}_q(t)$, our goal is to learn an equation of the form
\begin{equation}
    \dv{\boldsymbol{\pi}_q}{t} \equiv \frac{1}{T} \dv{\boldsymbol{\pi}_q}{\tau} = \mathcal F(\boldsymbol{\pi}_q; \boldsymbol{\pi}_p),
\end{equation}
where $T$ is a characteristic time scale, $\tau = t/T$ is the corresponding dimensionless time,  $\boldsymbol{\pi}_p$ is a vector of dimensionless parameters, and $\mathcal F$ is a differential operator. 
In the absence of a governing equation, SINDy approximates $\mathcal F$ by a sum of differential operators with fitting coefficients that are optimized for both sparsity and prediction error~\cite{brunton2016discovering}. That is, given input-output pairs $\left\{ \bm \pi_p, \bm \pi_q \right\}$ sampled from a time series, SINDy minimizes the loss
\begin{equation}\label{eq:sindy-loss}
    \mathcal L_\text{SINDy}(\bm \pi_q, \bm \pi_p, T ;\bm \Xi) = \left\| \frac{1}{T} \dv{\bm \pi_q}{\tau} - \bm \Theta\left(\bm \pi_q, \bm \pi_p ; T\right) \bm \Xi \right\|_2^2 + \lambda \left\| \bm \Xi \right\|_0,
\end{equation}
where $\bm \Xi$ contains unknown fitting coefficients and $\bm \Theta$ is a pre-determined library of potential candidate functions and differential operators, a linear combination of which makes up the approximation $\mathcal F = \bm \Theta\left(\bm \pi_q, \bm \pi_p\right) \bm \Xi$. For dimensional consistency, derivatives in the dependent variable $t$ are scaled by an unknown problem specific timescale $T$, which we assume can be expressed as a function of the input parameters: $T = T(\mathbf p)$.
In essence, the timescale $T$ is treated in the same manner as the dimensionless groups $\bm \pi$, except that its dimensions are constrained to be those of time.

Assuming that $\mathcal F(\boldsymbol{\pi}_q; \boldsymbol{\pi}_p)$, and its corresponding dictionary $\bm \Theta$, are \emph{separable}, we can write
\begin{equation}
\label{eq:sindy-ansatz}
    \bm \Theta(\boldsymbol{\bm \pi}_q; \boldsymbol{\pi}_p, T) = \bm g(\bm \pi_p) \otimes \hat{\bm{\Theta}}(\boldsymbol{\pi}_q, T),
\end{equation}
where $\hat{\bm{\Theta}}$ is a dictionary of $s$ derivatives in the dimensionless quantity of interest $\bm \pi_q$, $\bm \pi_p$ is a vector of dimensionless input parameter excluding time, $\bm g()$ is a dictionary of polynomial functions, and $\otimes$ is the Kronecker product which is a vectorization of the outer product of two vectors. 
The assumption of separability is not strictly necessary; candidate functions may be chosen that are functions of both $\bm \pi_p$ and $\bm \pi_q$.
However, SINDy libraries are often constructed as multinomials, so that the variables are separable.
Moreover, the parameters typically appear as coefficients of the state variable in both bifurcation normal forms and Taylor series approximations of dynamical systems, so the assumption of separability is natural.

The resulting feature space dimension is $\bm \Theta(\bm \pi_q; \bm \pi_p) \in {\mathbb R}^{ns}$. 
Having $m$ measurements, we define the full dictionary has dimensions $\bm{\Theta} \in {\mathbb R}^{m\times ns}$, where every row is an examples (i.e. a sample in time) and every column a potential term in the differential equation. Accordingly, $\bm \Xi \in {\mathbb R}^{ns \times k'}$ where $k'$ is the dimension of $\bm \pi_q$.

While equation \eqref{eq:sindy-loss} takes dimensionless pairs as input data, we would also like to optimize over candidate non-dimensionalizations. That is, we would like to minimize the following loss function

\begin{equation}\label{eq:dimless-sindy-loss}
    \mathcal L_\text{dSINDy}(\mathbf q, \mathbf p; \bm \Xi, \bm \Phi) = \mathcal L_\text{SINDy}\left( \bm \pi_p( \mathbf p; \bm \Phi), \bm \pi_q( \mathbf q; \bm \Phi), T(\mathbf p, \bm \Phi); \bm \Xi  \right),
\end{equation}
where the input-output pairs $(\mathbf p, \mathbf q)$ are sampled from the dimensional data, and $\bm \Phi$ is the Pi-groups powers matrix that defines the mapping $\bm \pi(\mathbf x; \bm \Phi) = \exp(\log(\mathbf x) \bm \Phi )$ given by equation~\eqref{eq:ptopi-mat}.

Equation \eqref{eq:dimless-sindy-loss} does not include a constraint on ${\bm \Phi}$ (Eq.~\eqref{eq:nulleqn}) to ensure that $\bm \pi$ is dimensionless. To address this issue, we generate a finite number of candidate dimensionless numbers up to a predetermined fractional power that satisfy equation~\eqref{eq:nulleqn}. The resulting set of non-dimensionalizations correspond to a set of power matrices $\bm {\bar \Phi} = \{\bm {\bar \Phi}_1, \bm {\bar \Phi}_2, \ldots, \bm {\bar \Phi}_r\}$ over which we minimize the loss
\begin{equation}
    \check{\bm{\Xi}}, \check{\bm{\Phi}} = \argmin_{\bm \Xi, i} \mathcal L_\text{dSINDy}(\mathbf p, \mathbf q; \bm \Xi, \bm{\bar \Phi}_i),
\end{equation}
with $i \in [1, r]$. $r$ depends on the range of predetermined powers from which dimensionless numbers are sampled according to the null-space condition in \eqref{eq:nulleqn}.

To avoid this ``brute-force`` search, the dimensionless SINDy method could be combined with the constrained optimization approach described in Sec.~\ref{sec:constrained-opt}.
In general, this tactic would be more flexible, since it also allows for non-integer powers in the dimensionless groups.
We use brute-force enumeration in this work because we expect integer powers, there are relatively few nullspace vectors, and the sparse regression problem can be solved efficiently, so this method scales relatively well in this particular case. We explore a more efficient approach in future work.
The general computational efficiency of the brute-force search approach will be discussed in the results.

In summary, this method solves two problems at once, providing
\begin{enumerate}
    \item Dimensionless input parameters $\boldsymbol{\pi}_p$ and dimensionless dependent variables $\tau = t/T$.
    \item A sparse and parametric dynamical system $\dot{\bm \pi}_q = \mathcal F(\bm \pi_q; \bm \pi_p)$.
\end{enumerate}
While we only consider time as a dependent variable in this section, a generalization to spatial and other dependent variables is straightforward. The method also allows for combining known dimensionless numbers with unknown ones as often encountered in practical problems.

\newpage
\section{Results}
In this section, we apply the three methods presented above on four non-dimensionalization problems: the harmonic oscillator in Sec.~\ref{sec:results-pendulum}, the bead on a rotating hoop in Sec.~\ref{sec:rotating-hoop}, the Blasius boundary layer in Sec.~\ref{sec:results-blasius}, and the Rayleigh-B\'enard problem in Sec.~\ref{sec:rayleigh-benard} . We discuss the advantages and shortcomings of each method in terms of accuracy, robustness, and speed in the context of our proposed hypothesis. 

\subsection{Non-dimensionalization as optimal change of variables: harmonic oscillator}
\label{sec:results-pendulum}
Dimensionless groups can act as scaling parameters (i.e. revealing self-similarity), as they often arise in analytical solutions. In some problems, Pi-groups can be deduced from an optimal change of variables without the need for a Buckingham Pi constraint. While this is not always the case, it supports the hypothesis that an optimal mapping between input and output data gives rise to a physically meaningful change of variables. We demonstrate this point on the simple pendulum problem described in Sec.\ref{sec:methods} and shown in Fig.~\ref{fig:approaches}. 

We evaluate the output prediction, $\alpha$, at 100 parameter combinations of the inputs $L$, $m$ and $g$ sampled from a uniform distribution, and at 100 time steps, $t$. In this problem, we employ the BuckiNet architecture \emph{without the Buckingham Pi constraint} (i.e. without $\mathcal L_\text{null}$) with four inputs $[L, m, g, t]$, one perceptron in the first hidden layer, a dimensionless output $\alpha$, 3 layers with 8 perceptrons each for $\psi$, and an exponential linear unit (ELU) activation function. The resulting dimensionless group in the BuckiNet layer is 
\begin{equation}
    \pi_p = \frac{gt^2}{L},
\end{equation}
which appears in the solution of the linear approximation of a harmonic oscillator as 
\begin{equation}
    \alpha(t) = \alpha_0\cos(\sqrt{\pi_p} + \theta),
\end{equation}
where $\theta$ is a constant that depends on the initial condition. This is a case where a change of variables that gives the simplest analytical solution is dimensionless. However, this is not generally the case, especially in higher dimensional problems. To fully take advantage of the Buckingham Pi constraint, we apply the three methods proposed in Sec.\ref{sec:methods} in the following examples.


\subsection{Bead on a rotating hoop}
\label{sec:rotating-hoop}
Consider a wire hoop with radius $R$, rotating about a vertical axis coinciding with its diameter at an angular velocity $\omega$, as shown in Fig.~\ref{fig:neural-nondim}. A bead with mass $m$ slides along the wire with tangential damping coefficient $b$. The equation governing the dynamics of the angular position $x$ of the bead with respect to the vertical axis is given by (\cite{strogatz2018nonlinear} sec. 3.5)
\begin{equation}\label{eq:rothoop}
    m R \ddot{x} = -b \dot{x} - m g \sin x + m R^2 \omega \sin x \cos x.
\end{equation}

A traditional dimensional analysis leads to the following Pi-groups~\cite{strogatz2018nonlinear}
\begin{equation}
\label{eq:hoop-numbers}
    \gamma = \frac{R \omega^2}{g}, \hspace{2cm} \epsilon = \frac{m^2 g R}{b^2},  \hspace{2cm} \tau = \frac{mg}{b} t,
\end{equation}
where $\epsilon$ controls the inertial term and $\gamma$ is a pitchfork bifurcation parameter that accounts for two additional fixed points at $x^* = \pm \arccos(\gamma)$, for $\gamma > 1$.
Non-dimensionlizing equation \eqref{eq:rothoop} gives
\begin{equation}
\label{eq:hoop-nondim}
    \epsilon \frac{d^2 x}{d \tau^2} = - \frac{d x}{d \tau} - \sin x + \gamma \sin x \cos x.
\end{equation}
For $\epsilon \ll 1$ and $\gamma = \mathcal{O}(1)$, the system is overdamped and approximately first-order.

\begin{table}[b]
\begin{center}
\begin{tabular}{ |c||c|c|c|c|c|} 
 \hline
 $\bm \Phi$ & $m$ & $R$ & $b$ & $g$ & $\omega$ \\ 
 \hline
 $\phi(\pi_1)$ & 0.0011 & 1.000 & 0.0001 & -0.997 & 1.990\\ 
 $\phi(\pi_2)$ & 1.991 & 1.000 & -1.990 & 0.998 & 0.002 \\
 \hline
\end{tabular}
\captionof{table}{Discovered Pi-groups powers with BuckiNet.}
\label{tab:buckinet-hoop-result}
\end{center}
\end{table} 

\begin{figure}[t]
\begin{center}
\includegraphics[width=\linewidth]{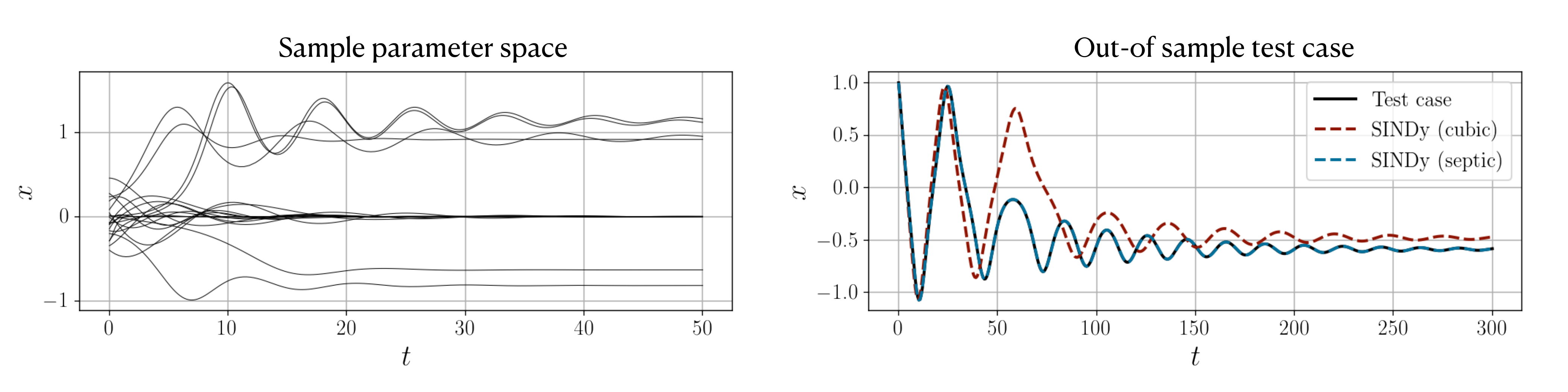}
\caption{Dimensionless SINDy applied on the rotating hoop problem. The discovered SINDy model reproduces the data and identifies the most physically relevant dimensionless groups.}
\label{fig:sindy-hoop}
\end{center}
\end{figure}

To test the BuckiNet and the constrained optimization algorithms, we solve the governing equations numerically to obtain 3000 solutions with mass, radius, damping coefficient, and angular velocity sampled from a uniform distribution. In order to recover $\gamma$ and $\epsilon$ without explicitly accounting for the time scale, we use the Principal Component Analysis of time series solution matrix (where each row corresponds to a different parameter combination, and each column to a different time sample), so that the output $\bm \pi_q$ is the set of coefficients for the leading $r$ principal components as shown in Fig.~\ref{fig:neural-nondim}. Here $\bm \pi_q \equiv \mathbf q = x$ because the angle $x$ is dimensionless. A sample of the time-series solutions for different parameters is shown in Fig.~\ref{fig:sindy-hoop}.

After optimizing over neural network hyperparameters -- given the optimal architecture, with enough modes to represent the solution -- BuckiNet recovers the correct dimensionless numbers shown in table \ref{tab:buckinet-hoop-result}. The discovered Pi-groups are scaled to make the power of $R$ unity.
In order to understand the limitations of the method, it is informative to look at sub-optimal solutions. For instance, for some hyperparameters (e.g. network architecture, regularization etc.) BuckiNet and the constrained optimization method give a product of either $\epsilon$ or $\gamma$ and a Pi-group that is closely related to $\tau$ as a solution, such that
\begin{equation}
    \phi(\pi_1) = \phi(\pi_2) + c \phi\left(\frac{mg}{\omega b}\right),
\end{equation}
where $c$ is an arbitrary constant, and $mg/(\omega b)$ is the closest approximation of $\tau=mgt/ b$, given that $t$ is not included as an input. This shows that a product of multiple Pi-groups is a local minimum that satisfies the Buckingham Pi and spans the solution space in a different coordinate system. However, a hyperparameter optimization over multiple trials will consistently result in Pi-groups that closely approximate $\gamma$ and $\epsilon$. This motivates the use of SINDy to further constrain the dimensionless numbers to be coefficients in a differential equation where they acquire a concrete mathematical meaning as controllers of the dominant balance.

Using the dimensionless SINDy approach, we generate 20 parameter combinations, sampled from a uniform distribution -- in this case, accounting for the time $t$ as an input. The library $\bm \Theta$ contains low-order polynomials in $x$ and $\dot{x}$ that can approximate the trigonometric nonlinearity in~\eqref{eq:rothoop} for relatively small values of $x$.
For each candidate nondimensionalization generated from the nullspace of $\bm D$, the nonconvex optimization problem~\eqref{eq:sindy-loss} is approximated with the simple sequentially thresholded least squares algorithm~\cite{brunton2016discovering}, where the only tuning parameter is a threshold that approximates the $\ell_0$ regularization loss.

\begin{figure}[t]
\vspace{-.35in}
\begin{center}
\includegraphics[width=.99\linewidth]{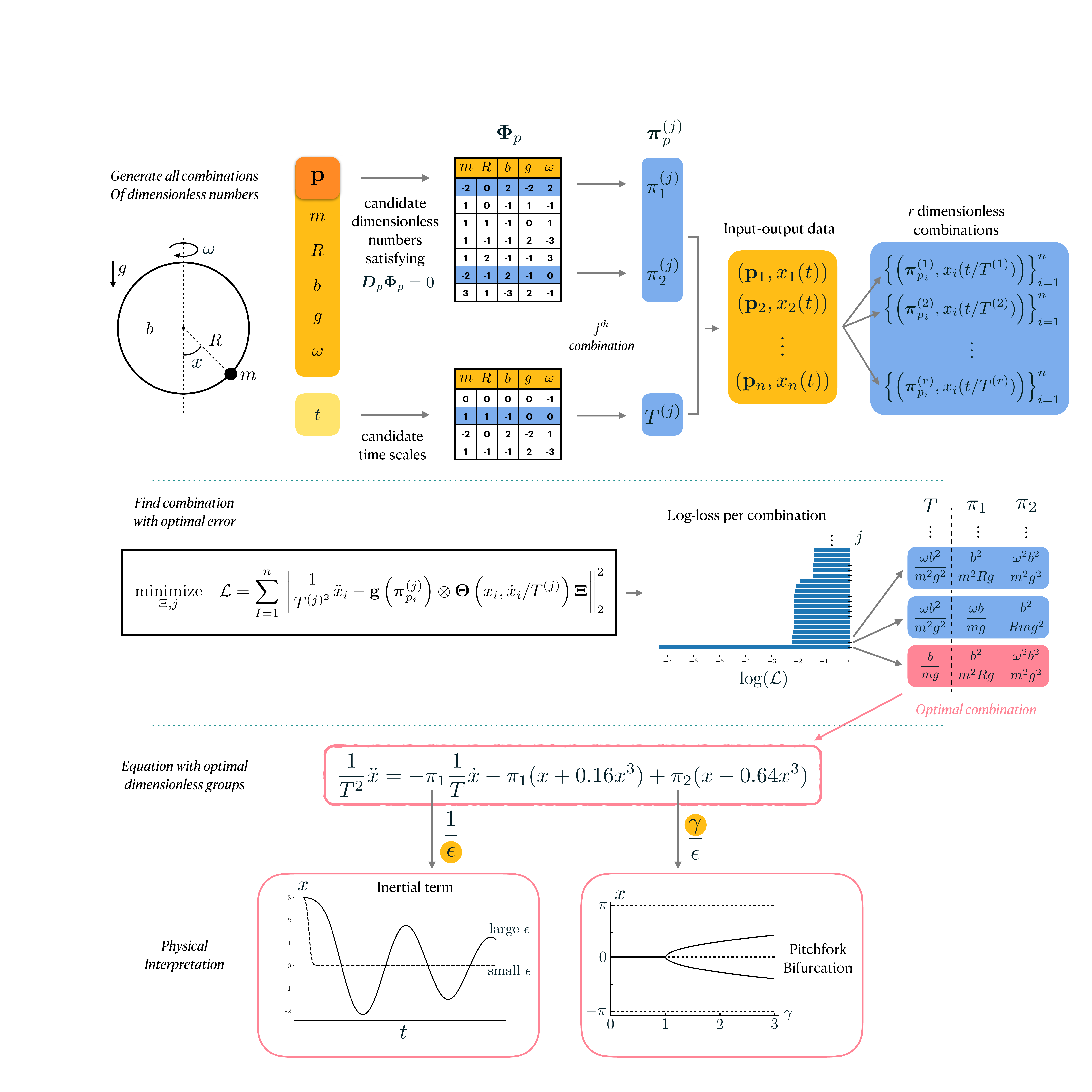}
\vspace{-.1in}
\caption{The dimensionless SINDy approach applied on the rotating hoop problem. i) generate candidate dimensionless numbers and time scales from the nullspace of the input powers matrix $\bm D_p$, ii) choose a combination of Pi-groups and a time scale, iii) cast as a SINDy problem with the chosen Pi-groups as coefficients, iv) choose the combination with the lowest SINDy loss.}
\label{fig:summary-figure}
\end{center}
\end{figure}
Fig.~\ref{fig:summary-figure} shows that the method identifies the same time scale $T = b/mg$ as that proposed by Strogatz~\cite{strogatz2018nonlinear} in equation~\eqref{eq:hoop-numbers}, along with ratios of $\epsilon$ and $\gamma$ which is consistent with equation~\eqref{eq:hoop-nondim} -- divided by $\epsilon$ -- such that
\begin{equation}
    \pi_1 = \frac{b^2}{R g m^2} = \frac{1}{\epsilon}, \hspace{2cm} \pi_2 = \frac{\omega^2 g^2}{m^2 g^2} = \frac{\gamma}{\epsilon}.
\end{equation}

Depending on the sparsity threshold in SINDy, models of varying fidelity can be identified.
For example, with a threshold of $10^{-1}$ the algorithm selects a cubic model
\begin{equation}
    \dv[2]{x}{\tau} = -0.96 \pi_1 \dv{x}{\tau} - 0.94 \pi_1 x +  \pi_2 (0.86x - 0.34x^3),
\end{equation}
while reducing the threshold to $10^{-3}$ results in the seventh-order model
\begin{equation}
    \dv[2]{x}{\tau} = - \pi_1 \dv{x}{\tau} -\pi_1(x - 0.16 x^3 + 0.01x^5) + \pi_2( x - 0.66 x^3 + 0.13 x^5 - 0.01 x^7).
\end{equation}
This closely matches the Taylor-series approximation of the true dynamics~\eqref{eq:hoop-nondim}.
Accordingly, an appropriate model can be selected based on a desired tradeoff between accuracy and simplicity. However, higher order models increase the accuracy of the discovered dimensionless groups.

Fig.~\ref{fig:sindy-hoop} tests the generalization of the two models on a set of parameters chosen to be outside the range of the training data (i.e. an extrapolation task).
Although the cubic model captures the qualitative behavior reasonably well, the seventh-order approximation closely tracks the true solution.

\subsection{Laminar boundary layer: identifying self-similarity}
\label{sec:results-blasius}

Non-dimensionalization often arises in the context of scaling and collapsing experimental results to lower dimensions by revealing the self-similarity structure of the solution space.
The discovery of self-similarity is considered an important result in many applications because it reveals universality properties.
There are plenty of such examples in the history of fluid dynamics, particularly in understanding turbulence~\cite{barenblatt1996scaling, Pope2000book}.

Boundary layer theory is an example where of nondimensionalization and self-similarity have played a central role in this understanding~\cite{Schlichting1955}.
Using scaling analysis, Prandtl showed that the Navier-Stokes equation describing an incompressible laminar boundary layer flow can be simplified to the \textit{boundary layer equations} in the streamfunction $\Psi$
\begin{equation}
\label{eq:boundary-layer-equations}
    \Psi_y \Psi_{xy} - \Psi_x \Psi_{yy} = \nu \Psi_{yyy},
\end{equation}
where the subscripts denote partial differentiation in $x$ and $y$, $\nu$ is the kinematic viscosity (with units of $L^2/T$). $\Psi$ is defined so that the streamwise and wall normal velocities are respectively given by
\begin{equation}
    u(x, y) = \Psi_y, \qquad v(x, y) = -\Psi_x.
\end{equation}

Although Prandtl's boundary layer equations are themselves a significant simplification and triumph of scaling analysis, they can be further simplified and expressed as an ordinary differential equation with the help of self-similarity.
Blasius took the scaling one step further by showing that if $\Psi(x, y)$ is a solution of~\eqref{eq:boundary-layer-equations}, then so is $\tilde{\Psi}(x, y) = \alpha \Psi(\alpha^2 x, \alpha y) $ for any dimensionless constant $\alpha$.
This in turn implies that $\Psi$ itself is not an independent function of $x$ and $y$, but of the combination $x/\sqrt{y}$.

Defining the dimensionless similarity variable $\eta$ and streamfunction $f = f(\eta)$ as
\begin{equation}
    \eta = y \sqrt{\frac{U_\infty}{\nu x} }, \qquad f(\eta) = \frac{\Psi(x, y)}{\sqrt{\nu U_\infty x}},
\end{equation}
with freestream velocity $U_\infty$, the boundary layer equations reduce to the nonlinear boundary value problem
\begin{subequations}
\label{eq:blasius-bvp}
    \begin{gather}
        f'''(\eta) + \frac{1}{2}f''(\eta) f(\eta) = 0,\\
        f(0) = f'(0) = 0, f'(\infty) = 1.
    \end{gather}
\end{subequations}
Although there is no known closed-form solution to this problem, it can be analyzed with asymptotic perturbation methods or solved numerically (e.g. with a shooting method to identify an appropriate initial value for $f''(0)$).

\begin{figure}[t]
\begin{center}
\includegraphics[width=16cm]{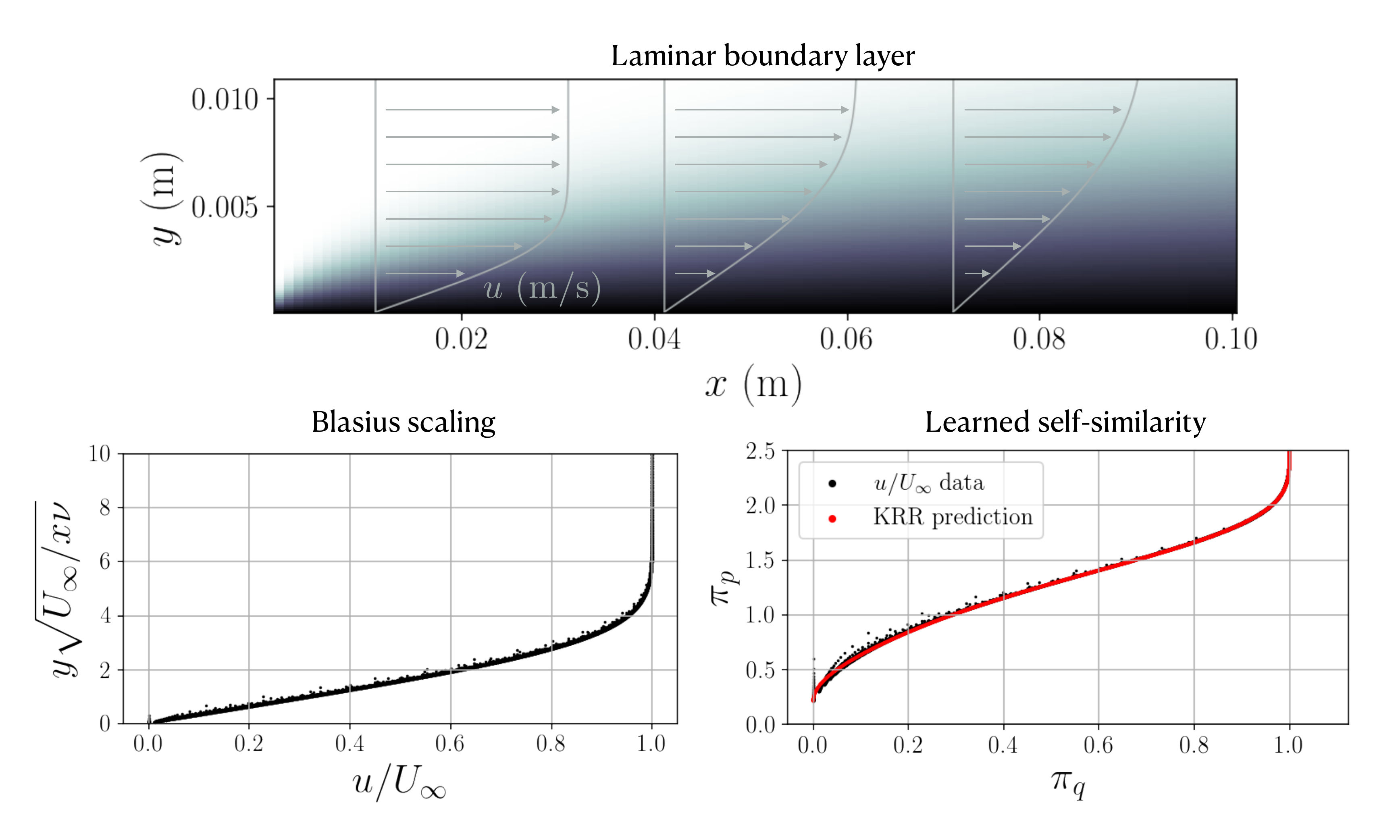}
\caption{Constrained optimization method for Blasius boundary layer problem, identifying the dimensionless group that collapses the input-output map to a single curve.}
\label{fig:krr-blasius}
\end{center}
\end{figure}

In contrast to the rotating hoop example, in this case the velocity profile is more important than the differential equation itself.
For instance, once the Blasius solution $f(\eta)$ is known, the boundary layer profile can be found by undoing the scaling, e.g. $u(x, y) = U_\infty f'(\eta)$.
The most natural method to directly identify a dimensionless group associated with the boundary layer is therefore the constrained optimization approach introduced in Sec.~\ref{sec:constrained-opt}.

Defining the output quantity of interest as the nondimensional streamwise velocity $\pi_q = u/U_\infty$, the problem is to learn a model for $\pi_q = \psi(\pi_p)$, where $\pi_p$ is an input dimensionless group that can depend on $x, y, U_\infty$, and $\nu$.
In this problem we seek to rediscover the Blasius scaling which was discovered analytically by Blasius and Prandtl with $\pi_p = \eta$ and $\psi = f'(\eta)$. 

We use kernel ridge regression (KRR) with a non-parametric radial basis function to approximate $\psi$.
Accordingly, the constrained optimization~\eqref{eq:constrained-opt} must learn a dimensionless group $\pi_p$ in the nullspace of the units matrix $\bm D$ that leads to a good KRR approximation of $\pi_q$ as a function of $\pi_q$.

We generate data for this example by solving the boundary layer equations~\eqref{eq:boundary-layer-equations} via a shooting method applied to~\eqref{eq:blasius-bvp}.
The free-stream velocity is chosen to be $U_\infty = 0.01 m/s$ with viscosity $\nu = 10^{-6} m^2/s$, close to that of water at room temperature.
The resulting two-dimensional profile $u(x, y)$ is shown in Fig.~\ref{fig:krr-blasius} (top), along with profiles at selected locations of $x$.
100 points in this field are selected randomly as training data and~\eqref{eq:constrained-opt} is solved with a constrained trust region method implemented in Scipy.

Specifically, for each candidate nondimensionalization, the kernel ridge regression model with radial basis function kernels (implemented in scikit-learn) is trained and evaluated on the dimensionless parameters computed from the 100 test points.
The KRR model uses ridge ($\ell_2$) penalty of $10^{-4}$, an $\ell_1$ penalty of $10^{-4}$, and a scale factor of $1$.
Note that in this method, generalizing on a test case is not crucial because the model is not used to make prediction but only as a proxy for the mutual information between the candidate nondimensionalization and the quantity of interest. In other words, the method does not have to address problems of overfitting.
Moreover, since only 100 points are used in training, the performance of the final model can be evaluated against the entire field ($10^4$ points).
The optimization problem is inexpensive but non-convex and sensitive to the initial guess, thus problem to converging to local minima. To address this issue, we run multiple optimizations with different initial guesses (around 20) and return the solution that has the minimum cost.

Fig.~\ref{fig:krr-blasius} compares the discovered nondimensionalization, $y^{0.46} U_\infty^{0.24}/(x^{0.22} \nu^{0.24}) \approx \sqrt{\eta}$, to the Blasius solution. When scaled to make the power of $y$ unity, the discovered dimensionless number is
\begin{equation}
\label{eq:blasius-result}
    \pi_p = \frac{y U_\infty^{0.51}}{x^{0.49} \nu^{0.51}} \approx \eta.
\end{equation}
The constrained optimization learns a different but equivalent model, in the sense that $\psi(\pi_p)$ is one-to-one with $f'(\eta)$.

In contrast to the brute force-type search over candidate nondimensionalizations generated from a search over vectors of integers, the exponents in this method are floating-point numbers and are arbitrary up to an overall constant.
The scale of the solution is therefore a balance between the $\ell_1$ and $\ell_2$ penalties and the scale factor in the radial basis functions. Setting the scale equal to one with small penalties thus biases the algorithm towards $\mathcal{O}(1)$ exponents.

\subsection{Rayleigh-B\'enard convection: learning a normal form}
\label{sec:rayleigh-benard}
Characterizing the onset and behavior of instabilities is crucial for understanding large-scale dynamical systems.
Near a critical point, a reduced-order amplitude equation called the normal form can model the qualitative change in dynamics associated with a bifurcation.
Although the normal form can be deduced analytically~\cite{GuckenheimerHolmes}, numerically~\cite{Meliga2009jfm, Carini2015jfm}, or via symmetry arguments~\cite{Golubitsky1988}, these methods become progressively more challenging with more complex systems.
With the exception of the symmetry analysis, they are also invasive because they require direct access to the (discretized) governing equations.
In this example, we use the SINDy method introduced in Sec.~\ref{sec:methods-sindy} to learn a dimensionless normal form from limited time-series data, along with the corresponding dimensional parameters.

\begin{figure}[t]
\begin{center}
\includegraphics[width=0.85\linewidth]{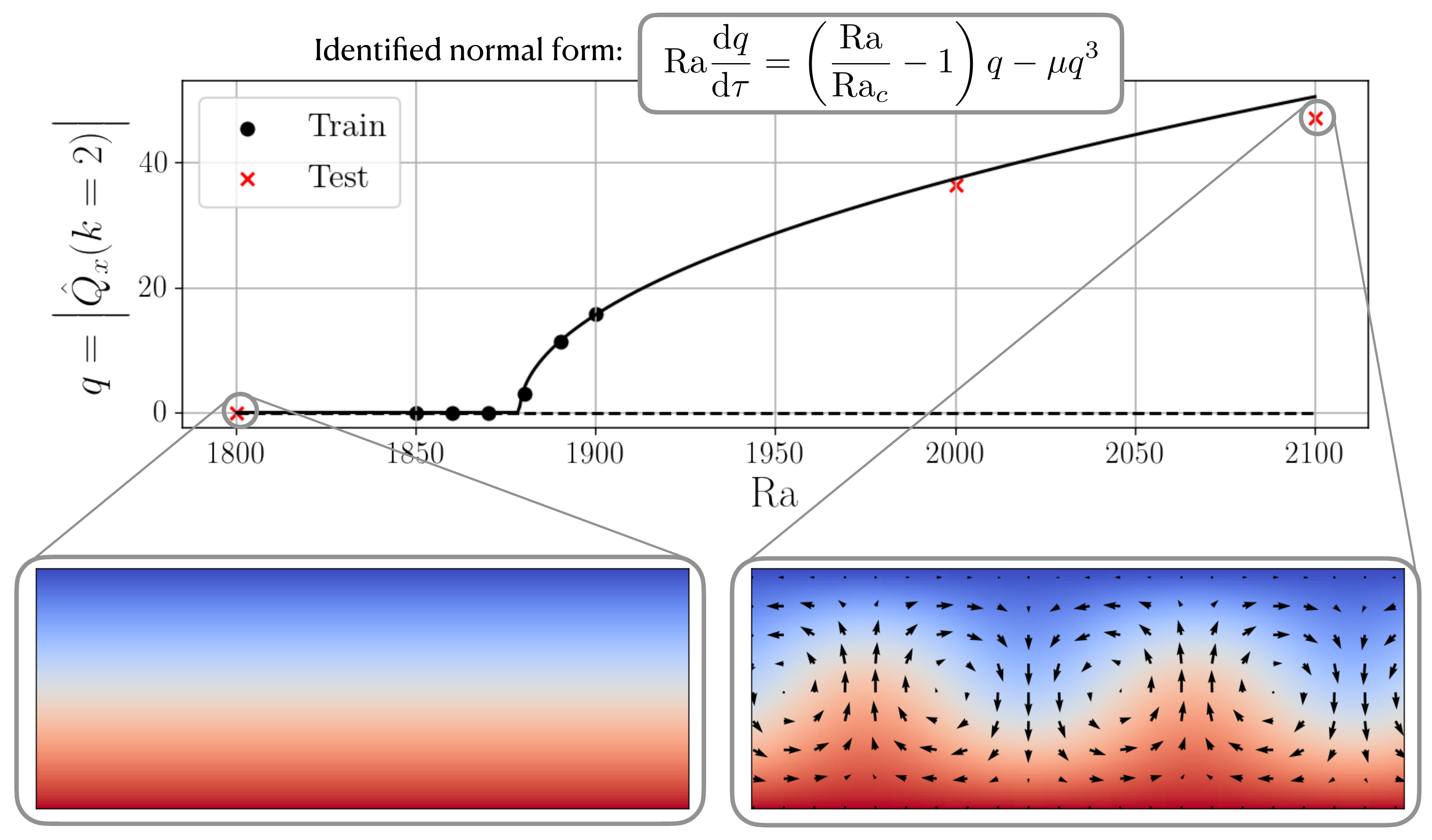}
\caption{The dimensionless SINDy method discovers the normal form of Rayleigh-B\'enard convection problem with the analytical form of the Rayleigh as a cofficient.}
\label{fig:sindy-RB}
\end{center}
\end{figure}

Rayleigh-B\'enard convection is a prototypical example of a system with a global instability that has been used in a range of studies of nonequilibrium dynamics, including pattern formation~\cite{Cross1993}, chaos~\cite{Lorenz1963jas}, and coherent structures in turbulence~\cite{Pandey2018ncomms}.
The system typically consists of a Boussinesq fluid between two plates, where the lower plate is at a higher temperature than the upper plate.
Heat transfer can take place via either conduction or convection, whose relative strength is encapsulated in the dimensionless Rayleigh number.
Below a critical Rayleigh number, the only stable solution is steady conduction between the plates; above it, the density gradient becomes unstable to convection, as shown in the lower panels of Fig.~\ref{fig:sindy-RB}.

For a fluid in the Boussinesq approximation linearized about density $\rho_0$ and temperature $T_0$, the governing equations consist of the conservation of momentum, mass, and energy:
\begin{subequations}
\begin{align}
    &\pdv{\mathbf{u}}{t} + (\mathbf{u} \cdot \nabla ) \mathbf{u} = -\rho_0^{-1} \nabla p + \nu \nabla^2 \mathbf{u} - g \alpha (T - T_0) \\
    &\nabla \cdot \mathbf{u} = 0 \\
    &\pdv{T}{t} + (\mathbf{u} \cdot \nabla) T = \kappa \nabla^2 T,
\end{align}
\end{subequations}
along with the boundary conditions that $T(z=0) = T_0$, $T(z=L_z) = T_0 + \Delta T$.
Besides the primitive variables, the system includes gravity $g$, coefficient of thermal expansion $\alpha$, kinematic viscosity $\nu$, and thermal diffusivity $\kappa$.
The flow is typically analyzed in terms of the Rayleigh number $\mathrm{Ra}=g\alpha \Delta T L_z^3 / \nu \kappa$ and Prandtl number $\mathrm{Pr}=\nu/\kappa$.
Although the Prandtl number has a significant impact on the behavior of the flow above the threshold of instability~\cite{Pandey2018ncomms}, the onset of instability itself is not sensitive to the Prandtl number~\cite{Chandrasekhar1961}.

We simulate the nondimensional flow in two dimensions with no-slip boundary conditions on the wall and periodic boundary conditions in the wall-parallel ($x$) direction using the Shenfun spectral Galerkin library~\cite{mortensen2018joss}.
We use a Fourier-Chebyshev discretization with $(N_x, N_z)=(256, 100)$ and choose $L_x = 2\pi L_z$.
The critical point in this simulation is at $\mathrm{Ra}_c \approx 1875$, significantly above the true value of $\mathrm{Ra}_c \approx 1708$.
This is likely due to differences in boundary conditions, wall-parallel domain extent, and the restriction to two dimensions.

We simulate the system at a range of Rayleigh numbers (shown in Fig.~\ref{fig:sindy-RB}) and $\mathrm{Pr}=0.7$.
For each simulation, we dimensionalize the data as follows.
We randomly select a value for $\kappa$ in the range $(0.5, 0.7) \mathrm{m}^2/\mathrm{s}$, $\Delta T$ in the range $(50, 80)~^\circ\mathrm{C} $, and $\alpha$ in the range $(1 \times 10^{-4}, 5 \times 10^{4}) \mathrm{m}/^{\circ}\mathrm{C}$. 
The separation $L_z$ is then computed for consistency with the simulation Rayleigh number, with $g = 9.8 \mathrm{m}/\mathrm{s}^2$.
Likewise, $\nu$ is computed for consistency with the Prandtl number.
Finally, time is dimensionalized based on a free-fall time scale $\sqrt{L_z/g\alpha \Delta T}$.
From eight initial simulation cases, we retain five points close to the bifurcation (black dots in Fig.~\ref{fig:sindy-RB}) to solve the dimensionless SINDy problem and withold the other three points (red crosses) as test data.

One option for learning a model of the flow near the critical point would be to input the spatiotemporally varying data and learn a PDE model~\cite{rudy2017data}.
This might yield something similar to a Ginzburg-Landau or Swift-Hohenberg model~\cite{Cross1993}.
Another approach is to approximate some quantity of interest with a linear modal decomposition such as proper orthogonal decomposition and fit an ODE model of the temporal coefficients~\cite{Loiseau2017jfm}.
Given the simple globally coherent spatial structure of the flow, we proceed with the latter.

To account for translational invariance, we first represent the temperature field with a Fourier series in the $x$-direction:
\begin{equation}
    T(x, z, t) = \sum_k \hat{T}_k(z, t) e^{ikx}.
\end{equation}
Recognizing that the weakly supercritical solution has only $k=\pm 2$ Fourier components (see Fig.~\ref{fig:sindy-RB}), we define a real-valued observable as the wall-normal integral of the negative temperature gradient (proportional to heat flux) at $k=2$:
\begin{equation}
    q(t) = \left| \int_0^{L_z}  -ik \hat{T}_2(z, t) \dd z \right|.
\end{equation}
Note that this choice is arbitrary; we could for instance model the coefficient of a particular Fourier-Chebyshev basis function, or that of a leading POD mode.
All would illustrate qualitatively similar behavior, but this definition of $q(t)$ is a simple and representative global observable.

We set up the dimensionless SINDy problem as described in Secs.~\ref{sec:methods-sindy}.
Since we expect a Landau-type dynamics model for the symmetry-breaking behavior, we choose a linear-cubic polynomial library in $q$ and search for one dimensionless group $\pi_p$, along with an appropriate dimensionless time $\tau$.
Since there are only six independent dimensional quantities in this problem $(H, g, \alpha, \Delta T, \nu, \kappa)$, it is inexpensive to perform the optimization over all dimensionless numbers composed of integer powers up to $\pm 2$.
This has the advantage of producing a simple number suitable for comparison to the classical result, but is not strictly necessary; the brute force search could be avoided with a constrained optimization approach as described in Sec.~\ref{sec:constrained-opt} and the boundary layer example in Sec.~\ref{sec:results-blasius}.

The optimal dimensionless number selected by the algorithm is the inverse of the Rayleigh number, $\pi_p = \mathrm{Ra}^{-1}$, with dimensionless time $\tau = t \kappa^2 / \alpha^2 \nu (\Delta T)^2 $.
The dynamics model is a normal form for a pitchfork bifurcation:
\begin{equation}
    \mathrm{Ra}\dv{q}{\tau} = \left(\frac{\mathrm{Ra}}{\mathrm{Ra}_c} - 1 \right) q - \mu q^3,
\end{equation}
with estimated critical Rayleigh number $\mathrm{Ra}_c \approx 1878$, growth rate $\lambda \approx 1.7 \times 10^{-3}$ and Landau parameter $\mu \approx 4.6\times 10^{-5}$.

The fixed points of this model can be found easily and compared to the steady states of the simulation, as shown in Fig.~\ref{fig:sindy-RB}.
The model closely matches the steady states not only of the training, but also of the withheld test points much farther from the bifurcation (shown as red crosses).
Of course, this is a simple model of a well-understood instability, but the ability to directly derive normal forms from dimensional data opens the possibility of modeling more complex bifurcations, including cases where the control parameters are not well established.

\section{Discussion}
Finally, we've shown that data can be used to discover an ``optimal'' set of dimensionless numbers that honor the Buckingham Pi theorem according to hypothesis \ref{main-hyp}.
However, the definition of that optimality is non-unique, leaving the open question of how to find the best way to integrate the data with the Buckingham Pi theorem.
In particular, the methods introduced in Sec.~\ref{sec:methods} all seek to identify dimensionless groups for which the corresponding function approximation (KRR, BuckiNet, or SINDy) can best represent the input-output relationship $\bm \pi_q = \psi(\bm \pi_p)$.
Choosing a SINDy representation of this relationship encodes an inductive bias towards low-order polynomials, which is often appropriate in dynamical systems applications or asymptotic expansions.
In these contexts, the identified parameters are more likely to have a familiar structure, such as the ratio of different terms in a governing equation.
On the other hand, the optimal dimensionless parameters as seen by more flexible and general representations such as neural networks or kernel regression may be completely different than those derived by classical analytic approaches giving insight into scaling Pi-groups that cannot be derived by hand.
For instance, the Blasius solution cannot be easily represented with low-order polynomials, but at the cost of interpretability of the function, kernel regression is able to accurately model the boundary layer with the nonstandard nondimensionalization in equation~\eqref{eq:blasius-result}.

More generally, dimensional analysis is a commonly used method for physics discovery and characterization.   Importantly, it highlights the fact that physical laws do not depend on any specific units of measurements.  Rather, physical laws are fundamental and have the property of {\em generalized homogeneity}~\cite{barenblatt1996scaling}, thus they are independent of the observer.  Advancements across scientific disciplines have used such concepts to develop governing equations,  from gravitation to electromagnetism.  Specifically, dimensionality reduction has identified critical symmetries, parametrizations, and bifurcation structures underlying a given model. As such, dimensionality reduction has always been a critical component of establishing the canonical models across disciplines.  However, when encountering new physics-based systems where the governing equations are unknown or their parametrizations undetermined, dimensionality reduction once again is critical for determining, {\em the general form to which any physical equation is reducible}~\cite{buckingham1914physically}.

Edgar Buckingham provided the rigorous mathematical foundations on which dimensional analysis is accomplished.  It is a constructive procedure which significantly constrains the space of hypothesized models.  Although greatly constraining the allowable model space, Buckingham Pi theory does not produce a unique model, but rather a small subset of parametrizations which are then typically chosen from expert knowledge or asymptotic considerations~\cite{barenblatt1996scaling}.  With modern machine learning, we have shown that the Buckingham Pi theory procedure can be automated to discovery a diversity of important physical features and properties.  In particular, we have shown that depending on the objective, there are three distinct methods by which we can produce a model through dimensional analysis.  Each method is framed as an optimization problem which either imposes hard, soft or no constraints on the null-space of the units matrix, or optimizes the robustness, accuracy and speed of the model selected. Specifically, we develop three data-driven techniques that use the Buckingham Pi theorem as a constraint: (i) a constrained optimization problem with a non-parametric input-output fitting function, (ii) a deep learning algorithm (BuckiNet) that projects the input parameter space to a lower dimension in the first layer, and (iii) a sparse identification of nonlinear dynamics (SINDy) based technique to discover dimensionless equations whose coefficients determine important features of the dynamics, such as inertia and bifurcations.  Such regularizations solve the ill-posed nature of Buckingham Pi theory and its non-unique solution space.  We demonstrated the application of each method on a set of well-known physics problems, showing that without any prior knowledge of the underlying physics, the various architectures are able to extract all the key concepts from the data alone. 

The suite of algorithms introduced provide a set of physics-informed learning tools that can help characterize new systems and the underlying general form to which the physical system is reducible, regardless of whether governing equations are known or not.  This includes extracting in an automated fashion, the system's symmetries, parametric dependencies and potential bifurcation parameters.  Although modern machine learning can simply learn accurate representations of input and output relations, the imposition of Buckingham Pi theory allows for interpretable or explainable models.   Explainable AI, especially in the context of physics-based systems, has grown in importance as it is imperative to understand the feature space on which modern AI empowered autonomy, for instance, makes decisions with guaranteed safety margins.

\section*{Acknowledgments}
The authors acknowledge support from the Army Research Office (ARO W911NF-19-1-0045) and the National Science Foundation AI Institute in Dynamic Systems (Grant No. 2112085).
JLC acknowledges funding support from the Department of Defense (DoD) through the National Defense Science \& Engineering Graduate (NDSEG) Fellowship Program.


 \begin{spacing}{.9}
 \small{
 \setlength{\bibsep}{5.5pt}
 \bibliographystyle{unsrt}
 \bibliography{references}

\begin{thebibliography}{10}

\bibitem{barenblatt1996scaling}
Grigory~Isaakovich Barenblatt.
\newblock {\em Scaling, self-similarity, and intermediate asymptotics:
  dimensional analysis and intermediate asymptotics}.
\newblock Number~14. Cambridge University Press, 1996.

\bibitem{sterrett2017physically}
Susan~G Sterrett.
\newblock Physically similar systems-a history of the concept.
\newblock In {\em Springer handbook of model-based science}, pages 377--411.
  Springer, 2017.

\bibitem{buckingham1914physically}
Edgar Buckingham.
\newblock On physically similar systems; illustrations of the use of
  dimensional equations.
\newblock {\em Physical review}, 4(4):345, 1914.

\bibitem{del2019lurking}
Zachary del Rosario, Minyong Lee, and Gianluca Iaccarino.
\newblock Lurking variable detection via dimensional analysis.
\newblock {\em SIAM/ASA Journal on Uncertainty Quantification}, 7(1):232--259,
  2019.

\bibitem{jofre2020data}
Llu{\'\i}s Jofre, Zachary~R del Rosario, and Gianluca Iaccarino.
\newblock Data-driven dimensional analysis of heat transfer in irradiated
  particle-laden turbulent flow.
\newblock {\em International Journal of Multiphase Flow}, 125:103198, 2020.

\bibitem{fukami2021robust}
Kai Fukami and Kunihiko Taira.
\newblock Robust machine learning of turbulence through generalized buckingham
  pi-inspired pre-processing of training data.
\newblock In {\em APS Division of Fluid Dynamics Meeting Abstracts}, pages
  A31--004, 2021.

\bibitem{xie2021data}
Xiaoyu Xie, Wing~Kam Liu, and Zhengtao Gan.
\newblock Data-driven discovery of dimensionless numbers and scaling laws from
  experimental measurements.
\newblock {\em arXiv preprint arXiv:2111.03583}, 2021.

\bibitem{Cerda2003prl}
E.~Cerda and L.~Mahadevan.
\newblock Geometry and physics of wrinkling.
\newblock {\em Phys. Rev. Letters}, 90(7):074302, 2003.

\bibitem{Morris1993prl}
S.~W. Morris, E.~Bodenschatz, D.~S. Cannell, and G.~Ahlers.
\newblock Spiral defect chaos in large aspect ratio {Rayleigh-B{\`e}nard}
  convection.
\newblock {\em Physical Review Letters}, 71(13):2026, 1993.

\bibitem{Shi1994science}
X.~D. Shi, M.~P. Brenner, and S.~R. Nagel.
\newblock A cascade of structure in a drop falling from a faucet.
\newblock {\em Science}, 265(5169):219--222, 1994.

\bibitem{Grzybowski2000nature}
B.~Grzybowski, H.~A. Stone, and G.~M. Whitesides.
\newblock Dynamic self-assembly of magnetized, millimetre-sized objects
  rotating at a liquid-air interface.
\newblock {\em Nature}, 405:1033--1036, 2000.

\bibitem{Seminara2012pnas}
A.~Seminara, T.~E. Angelini, J.~N. Wilking, H.~Vlamakis, S.~Ebrahim, R.~Kolter,
  D.~A. Weitz, and M.~P. Brenner.
\newblock Osmotic spreading of bacillus subtilis biofilms driven by an
  extracellular matrix.
\newblock {\em Proceedings of the National Academy of Sciences},
  109(4):1116--1121, 2012.

\bibitem{Cross1993}
M.~C. Cross and P.~C. Hohenberg.
\newblock Pattern formation outside of equilibrium.
\newblock {\em Reviews of modern physics}, 65(3):851, 1993.

\bibitem{callaham2021learning}
Jared~L Callaham, James~V Koch, Bingni~W Brunton, J~Nathan Kutz, and Steven~L
  Brunton.
\newblock Learning dominant physical processes with data-driven balance models.
\newblock {\em Nature communications}, 12(1):1--10, 2021.

\bibitem{guckenheimer_holmes}
Philip Holmes and John Guckenheimer.
\newblock {\em Nonlinear oscillations, dynamical systems, and bifurcations of
  vector fields}, volume~42 of {\em Applied Mathematical Sciences}.
\newblock Springer-Verlag, Berlin, Heidelberg, 1983.

\bibitem{Yair2017pnas}
Or~Yair, Ronen Talmon, Ronald~R Coifman, and Ioannis~G Kevrekidis.
\newblock Reconstruction of normal forms by learning informed observation
  geometries from data.
\newblock {\em Proceedings of the National Academy of Sciences}, page
  201620045, 2017.

\bibitem{kalia2021learning}
Manu Kalia, Steven~L Brunton, Hil~GE Meijer, Christoph Brune, and J~Nathan
  Kutz.
\newblock Learning normal form autoencoders for data-driven discovery of
  universal, parameter-dependent governing equations.
\newblock {\em arXiv preprint arXiv:2106.05102}, 2021.

\bibitem{schmidt2009distilling}
Michael Schmidt and Hod Lipson.
\newblock Distilling free-form natural laws from experimental data.
\newblock {\em science}, 324(5923):81--85, 2009.

\bibitem{brunton2016discovering}
Steven~L Brunton, Joshua~L Proctor, and J~Nathan Kutz.
\newblock Discovering governing equations from data by sparse identification of
  nonlinear dynamical systems.
\newblock {\em Proceedings of the national academy of sciences},
  113(15):3932--3937, 2016.

\bibitem{rudy2017data}
Samuel~H Rudy, Steven~L Brunton, Joshua~L Proctor, and J~Nathan Kutz.
\newblock Data-driven discovery of partial differential equations.
\newblock {\em Science Advances}, 3(4):e1602614, 2017.

\bibitem{lu2021learning}
Lu~Lu, Pengzhan Jin, Guofei Pang, Zhongqiang Zhang, and George~Em Karniadakis.
\newblock Learning nonlinear operators via deeponet based on the universal
  approximation theorem of operators.
\newblock {\em Nature Machine Intelligence}, 3(3):218--229, 2021.

\bibitem{raissi2019physics}
Maziar Raissi, Paris Perdikaris, and George~E Karniadakis.
\newblock Physics-informed neural networks: A deep learning framework for
  solving forward and inverse problems involving nonlinear partial differential
  equations.
\newblock {\em Journal of Computational Physics}, 378:686--707, 2019.

\bibitem{karniadakis2021physics}
George~Em Karniadakis, Ioannis~G Kevrekidis, Lu~Lu, Paris Perdikaris, Sifan
  Wang, and Liu Yang.
\newblock Physics-informed machine learning.
\newblock {\em Nature Reviews Physics}, 3(6):422--440, 2021.

\bibitem{Noe2019science}
Frank No{\'e}, Simon Olsson, Jonas K{\"o}hler, and Hao Wu.
\newblock Boltzmann generators: Sampling equilibrium states of many-body
  systems with deep learning.
\newblock {\em Science}, 365(6457):eaaw1147, 2019.

\bibitem{Brenner2019prf}
MP~Brenner, JD~Eldredge, and JB~Freund.
\newblock Perspective on machine learning for advancing fluid mechanics.
\newblock {\em Physical Review Fluids}, 4(10):100501, 2019.

\bibitem{Duraisamy2019arfm}
Karthik Duraisamy, Gianluca Iaccarino, and Heng Xiao.
\newblock Turbulence modeling in the age of data.
\newblock {\em Annual Reviews of Fluid Mechanics}, 51:357--377, 2019.

\bibitem{Brunton2020arfm}
Steven~L. Brunton, Bernd~R. Noack, and Petros Koumoutsakos.
\newblock Machine learning for fluid mechanics.
\newblock {\em Annual Review of Fluid Mechanics}, 52:477--508, 2020.

\bibitem{Sonnewald2021}
M.~Sonnewald, R.~Lguensat, D.~C. Jones, P.~D. Dueben, J.~Brajard, and
  V.~Balaji.
\newblock Bridging observations, theory, and numerical simulation of the ocean
  using machine learning.
\newblock {\em Environmental Research Letters}, 16(7):073008, 2021.

\bibitem{Kaiser2021}
B.~E. Kaiser, J.~A. Saenz, M.~Sonnewald, and D.~Livescu.
\newblock Objective discovery of dominant dynamical processes with intelligible
  machine learning.
\newblock {\em arXiv:2106.12963}, 2021.

\bibitem{Sonnewald2019ess}
M.~Sonnewald, C.~Wunsch, and P.~Heimbach.
\newblock Unsupervised learning reveals geography of global ocean regimes.
\newblock {\em Earth and Space Science}, 6:784--794, 2019.

\bibitem{wu2018deep}
Hao Wu, Andreas Mardt, Luca Pasquali, and Frank Noe.
\newblock Deep generative markov state models.
\newblock {\em 32nd Conference on Neural Information Processing Systems
  (NeurIPS)}, 2018.

\bibitem{Champion2019pnas}
K.~Champion, B.~Lusch, J.~Nathan Kutz, and Steven~L. Brunton.
\newblock Data-driven discovery of coordinates and governing equations.
\newblock {\em Proceedings of the National Academy of Sciences},
  116(45):22445--22451, 2019.

\bibitem{bakarji2022discovering}
Joseph Bakarji, Kathleen Champion, J~Nathan Kutz, and Steven~L Brunton.
\newblock Discovering governing equations from partial measurements with deep
  delay autoencoders.
\newblock {\em arXiv preprint arXiv:2201.05136}, 2022.

\bibitem{Brunton2019book}
S.~L. Brunton and J.~N. Kutz.
\newblock {\em Data-Driven Science and Engineering: Machine Learning, Dynamical
  Systems, and Control}.
\newblock Cambridge University Press, 2019.

\bibitem{constantine2017data}
Paul~G Constantine, Zachary del Rosario, and Gianluca Iaccarino.
\newblock Data-driven dimensional analysis: algorithms for unique and relevant
  dimensionless groups.
\newblock {\em arXiv preprint arXiv:1708.04303}, 2017.

\bibitem{udrescu2020ai}
Silviu-Marian Udrescu and Max Tegmark.
\newblock Ai feynman: A physics-inspired method for symbolic regression.
\newblock {\em Science Advances}, 6(16):eaay2631, 2020.

\bibitem{gunaratnam2003improving}
David~J Gunaratnam, Taivas Degroff, and John~S Gero.
\newblock Improving neural network models of physical systems through
  dimensional analysis.
\newblock {\em Applied Soft Computing}, 2(4):283--296, 2003.

\bibitem{strogatz2018nonlinear}
Steven~H Strogatz.
\newblock {\em Nonlinear dynamics and chaos: With applications to physics,
  biology, chemistry, and engineering}.
\newblock CRC press, 2018.

\bibitem{Pope2000book}
S.~B. Pope.
\newblock {\em Turbulent Flows}.
\newblock Cambridge University Press, 2000.

\bibitem{Schlichting1955}
H.~Schlichting.
\newblock {\em Boundary-Layer Theory}.
\newblock McGraw-Hill, 1955.

\bibitem{GuckenheimerHolmes}
J.~Guckenheimer and P.~Holmes.
\newblock {\em Nonlinear oscillations, dynamical systems, and bifurcations of
  vector fields}.
\newblock Springer, 1983.

\bibitem{Meliga2009jfm}
P.~Meliga, J.-M. Chomaz, and D.~Sipp.
\newblock Global mode interaction and pattern selection in the wake of a disk:
  a weakly nonlinear expansion.
\newblock {\em Journal of Fluid Mechanics}, 633:159--189, 2009.

\bibitem{Carini2015jfm}
M.~Carini, F.~Auteri, and F.~Giannetti.
\newblock Centre-manifold reduction of bifurcating flows.
\newblock {\em J. Fluid Mech.}, 767:109--145, feb 2015.

\bibitem{Golubitsky1988}
M.~Golubitsky and W.~Langford.
\newblock Pattern formation and bistability in flow between counterrotating
  cylinders.
\newblock {\em Physica D}, 32:362--392, 1988.

\bibitem{Lorenz1963jas}
E.~N. Lorenz.
\newblock Deterministic nonperiodic flow.
\newblock {\em Journal of the Atmospheric Sciences}, 20(2):130--141, Mar 1963.

\bibitem{Pandey2018ncomms}
A.~Pandey, J.~D. Scheel, and J.~Schumacher.
\newblock Turbulent superstructures in {Rayleigh-B\`enard} convection.
\newblock {\em Nature Communications}, 9(2118), 2018.

\bibitem{Chandrasekhar1961}
S.~Chandrasekhar.
\newblock {\em Hydrodynamic and Hydromagnetic Stability}.
\newblock Clarendon Press, Oxford, 1961.

\bibitem{mortensen2018joss}
Mikael Mortensen.
\newblock Shenfun: High performance spectral galerkin computing platform.
\newblock {\em Journal of Open Source Software}, 3(31):1071, 2018.

\bibitem{Loiseau2017jfm}
J.-C. Loiseau and S.~L. Brunton.
\newblock Constrained sparse {Galerkin} regression.
\newblock {\em Journal of Fluid Mechanics}, 838:42--67, 2018.

\end{thebibliography}
 }
 \end{spacing}

\end{document}